\documentclass{article}
\usepackage{spconf,amsmath,graphicx}
\DeclareMathOperator*{\argmaxA}{arg\,max}
\newcommand{\ngram}[0]{G}
\newcommand{\minus}{\scalebox{0.5}[1.0]{$-$}}

\title{A likelihood ratio based domain adaptation method for E2E models}
\name{Chhavi Choudhury, Ankur Gandhe, Xiaohan Ding, Ivan Bulyko}
\address{Amazon Alexa}
\begin{document}
%
\maketitle
\begin{abstract}
End-to-end (E2E) automatic speech recognition models like Recurrent Neural Networks Transducer (RNN-T) are becoming a popular choice for streaming ASR applications like voice assistants. While E2E models are very effective at learning representation of the training data they are trained on, their accuracy on unseen domains remains a challenging problem. Additionally, these models require paired audio and text training data, are computationally expensive and are difficult to adapt towards the fast evolving nature of conversational speech. In this work, we explore a contextual biasing approach using likelihood-ratio that leverages text data sources to adapt RNN-T model to new domains and entities. We show that this method is effective in improving rare words recognition, and results in a relative improvement of 10\% in 1-best word error rate (WER) and 10\% in n-best Oracle\footnote{n-best Oracle WER is measured by scanning the n-best hypotheses produced from the beam search and selecting the lowest word-error hypothesis to compute WER} WER (n=8) on multiple out-of-domain datasets without any degradation on a general dataset. We also show that complementing the contextual biasing adaptation with adaptation of a second-pass rescoring model gives additive WER improvements.
\end{abstract}

\begin{keywords}
RNN-T, contextual biasing, shallow fusion, rare words, adaptation
\end{keywords}
\section{Introduction}
\label{sec:intro}
Hybrid Automatic Speech Recognition (ASR) models consist of separately trained models for acoustics, pronunciations and  language, \cite{DNN_SM, li2020developing}, whereas end-to-end (E2E) models integrate these components into a single network \cite{graves2012sequence, graves2013speech, 2019TaraMobile}, enabling end-to-end training and optimization. Latest advancements in the ASR technology have popularized E2E models like RNN-T as they provide state-of-the-art performance across a wide variety of streaming applications \cite{graves2013speech}. At the same time, the benefit of hybrid models is that they can take advantage of a variety of data sources, especially large amounts of text-only data and pronunciation lexicons, whereas E2E models need to be trained with paired audio and text data which can be expensive to collect and therefore limited.

Various techniques have been proposed to adapt E2E models to out-of-domain (OOD) data using text-only data sources during inference to bias the model predictions. For example, use of language models like n-gram Language Models (LMs)  \cite{chorowski2016towards, bahdanau2016end} or RNNs \cite{kannan2018analysis, peyser2020improving} in shallow fusion to incorporate external text corpora to improve the overall performance of the model has already been extensively studied. \cite{narayanan2019recognizing} observes that end-to-end models are particularly sensitive to domain-mismatch between training and inference, caused by overfitting to the training domain. OOD data usually contains words and phrases that RNN-T has not seen before, resulting in poor accuracy for these rare words.

Our work is inspired by \cite{9003790} which uses Bayes' rule to obtain RNN-T posteriors by using the language model probability $p(w)$ from a separate LM. \cite{9383515} also suggests that the internal ``LM" learnt by the RNN-T model needs to be subtracted for domain adaptation. Unlike previous work which looked at complete domain shift at test time, our objective is to adapt the RNN-T model on newer entities and domains as the data evolves, without causing degradation on already supported domains. To the best of our knowledge, there is no published work that uses n-gram model based likelihood ratio for domain adaptation. In this paper, we build a non-stochastic boosting model from an n-gram LM that can adapt the RNN-T model to large number of domains in a scalable way. We also show that the improvement from first pass (FP) shallow fusion (SF) is additive to the improvements from a second pass rescoring model. 

\section{Background}
\label{sec:background}

The RNN-T model decodes incoming audio signals directly to transcripts for downstream tasks. The objective function for decoding a sequence-to-sequence model typically would be given:

\begin{equation}
\label{eq:eq1}
\hat{y} = \argmaxA_y log\ p(y|x)
\end{equation}

where $x$ is the audio signal, $y$ is the output word sequence and $p(y|x)$ is the probability generated by the RNN-T model. A typical way to incorporate an external LM is using SF, which can be seen as a mixture of expert models: 
\begin{equation}
\label{eq:eq_sf}
\hat{y} = \argmaxA_y (\log\ p(y|x)  + \lambda\ \log p_{LM}(y) )
\end{equation}

where $p_{LM}$ is provided by an external language model and $\lambda\ $ is a tunable parameter.

The LM model can be an n-gram LM or a neural LM (NLM) like RNN. Shallow fusion is a very popular choice for incorporating text based LMs in ASR as they achieve significant performance gains \cite{chorowski2016towards, kannan2018analysis, zeyer2018improved,  battenberg2017exploring, Zhao2019ShallowFusionEC}. The language model is trained separately on a text data source. The LM scores are log-linearly interpolated with the RNN-T scores at each step of the beam search during inference. The interpolation weight $\lambda$ is determined by running a weight sweep and subsequent evaluation against a target development set and control set. The hypotheses from the first pass decoding can be further rescored during second pass using other LMs, so that the final hypothesis is selected using:

\begin{equation}
\label{eq:eq_sp}
\begin{aligned}
\hat{y} = \argmaxA_y ( log\ P(y|x) + \lambda\ log\ P_{LM}(y) \\ + \alpha\ log P_{RLM}(y)
\end{aligned}
\end{equation}

\begin{table*}[]
\begin{tabular}{|l|llllll|llll|}
\hline
              & \multicolumn{6}{c|}{Example 1: LM Score for rare OOD utterance}                   & \multicolumn{4}{c|}{Example 2: LM Score for general traffic utterance} \\ \cline{2-11} 
              & tune  & into  & the   & freiberg & game  & \textless{}/s\textgreater{} & play      & some      & music     & \textless{}/s\textgreater{}    \\ \hline
log $P_{GEN}$ & -8.12 & -2.86 & -2.55 & -15.64   & -7.38 & -1.63                       & -2.32     & -4        & -1.46     & -0.32                          \\
log $P_{OOD}$ & -9.37 & -5.55 & -2.74 & -6.87    & -4.94 & -1.9                        & -3        & -5.23     & -3.79     & -1.83                          \\
LLR           & -1.24 & -2.69 & -0.19 & 8.77     & 2.45  & -0.27                       & -0.69     & -1.23     & -2.33     & -1.5                           \\
Boost         & 0     & 0     & 0     & 8.77     & 0     & 0                           & 0         & 0         & 0         & 0                              \\ \hline
\end{tabular}
\caption{Examples: LM Scores for Rare and Common utterance}
\label{examples}
\end{table*}

\section{Adaptation to domain data}
Our objective is to adapt a pre-trained RNN-T model on a given OOD dataset $D_{OOD}$ without degrading the performance on domains already supported by the data used to train the RNN-T model. Hence, unlike \cite{9383515}, we do not want to completely remove the internal LM scores learnt by the model. Instead, we propose to use a non-stochastic boosting approach that learns to boost n-grams that are out-of-domain w.r.t to the RNN-T model training data without impacting the general domains.

\subsection{N-gram selection}
\label{sec:methodology}
 
Our n-gram selection algorithm is based on entropy difference between the distribution $P_{GEN}$ of the the general RNN-T dataset and distribution $P_{OOD}$ of the out-of-domain dataset. It has been shown in ~\cite{mezzoudj2018textual} that for select in-domain data from an out of domain corpus, using the entropy difference $H_{P_{GEN}}(s) - H_{P_{OOD}}(s)$ for ranking candidate sentences $s \in S$ is an effective strategy. Similarly, for selecting n-grams that need to be boosted during shallow fusion, we use the difference in likelihood for detecting ``surprising" n-grams in the out of domain dataset.    
 
More concretely, we build an n-gram statistical language model (SLM) using OOD data to estimate $P_{OOD}$. This is the data distribution that the RNN-T model has not seen before. We also build an n-gram SLM using the data used for training the RNN-T model to estimate $P_{GEN}$, the data distribution that the RNN-T model is trained for. The rare phrases present in OOD training data will have higher likelihood in $P_{OOD}$ and lower likelihood in $P_{GEN}$. We obtain a score for each n-gram by generating a log-likelihood ratio score for every n-gram $\ngram$ in the $P_{GEN}$ model as:
\begin{equation}
\begin{aligned}
LLR(\ngram) = log(p_{OOD}(\ngram)) - log(p_{GEN}(\ngram))
\end{aligned}
\end{equation}
such that in the set $S$, the rare words phrases will have positive log likelihood. We use the score $LLR$ of every n-gram as the boosting weight when $LLR(\ngram) > 0$. 


\subsection{N-gram pruning}
Boosting every n-gram which has a positive likelihood ratio score would result in boosting of a large number of n-grams, many of which would not be rare in the RNN-T dataset. To be more selective, we boost only when the difference in likelihood is greater than a certain threshold $T$:  
\begin{equation}
    S(\ngram):=
     \begin{cases}
       LLR(\ngram) & \text{if}\ LLR(\ngram) > T \\
       0, & \text{otherwise},
     \end{cases}
\end{equation}

The threshold is determined empirically by sweeping various combinations of threshold values and language model weights in the first pass. The higher the threshold, the more selective we are about boosting words. We implemented this approach by building a WFST (Weighted Finite State Transducer) from the n-gram model and setting the arc costs lower than the threshold $T$ to 0. During inference, we use shallow fusion to incorporate this WFST in the RNN-T model.

 


We show how this method works using some example utterances. In Table \ref{examples} Example 1 , the first utterance from OOD training data is ``tune into the freiburg game". The phrase ``freiburg game", is a rare phrase in general dataset. It gets a much smaller LM likelihood from $P_{OOD}$  $(\minus6.87 + \minus4.94= \minus11.81)$ than $P_{GEN}$  $(\minus15.64 + \minus7.38 = \minus23.02)$. This results in high log-likelihood ratio scores in the the subtracted model (8.77 + 2.45 = 11.22) for this phrase. After clipping the scores smaller  than threshold $3$, we get the score for the rare word ``freiburg" ($8.77$) which will help in boosting this word when seen during RNN-T beam search. Note that in the final WFST, we keep the arcs corresponding to the context so that the word ``freiburg" is boosted only in the relevant context. 

Table \ref{examples}  Example 2 shows how the scores shape up for a common utterance from general dataset ``play some music". The $P_{GEN} $ has a higher likelihood for the utterance as compared to $P_{OOD}$. Therefore, all likelihood ratio scores are negative for this utterance, and after using a threshold of $3$, the total boosting score comes out be 0. Hence, the boosting FST will only selectively boost rare phrases when applied using shallow fusion to RNN-T model.

In practice, we want to adapt the model to multiple OOD datasets representing various domains and new feature requests. When building the $P_{OOD}$, we build a separate n-gram model for each domain and linearly interpolate them with equal weights to get the final OOD distribution to use for n-gram selection and pruning. This helps normalize the variation in training dataset sizes. We leave the effect of optimizing interpolation weights for best overall perplexity of OOD data as future work.


\section{Experimental Setup}
\label{sec:expt_setup}

Following \cite{guo2020efficient}, our baseline RNN-T model consists of an encoder comprised of five LSTM layers of size 1024, and a two-layer LSTM prediction network of size 1024 with an embedding layer of 512 units. The softmax layer consists of 4k (subword) output units. The training data consists of over 200k hours of de-identified\footnote{The data was processed such that the users are not identifiable} utterances from interactions with Alexa voice assistant. The model was trained first with the RNN-T loss and then with minimum word error rate (MWER) criterion as described in ~\cite{guo2020efficient,shannon2017optimizing}. For shallow fusion with a WFST, we use the lookahead approach described in ~\cite{gourav2021personalization} as it avoids unnecessary arc expansion and provides a heuristic approach to perform subword-level rescoring without the need to build the boosting FST directly at the subword level. To optimize the search for arcs that have a common prefix string, we sort the input arc in lexicographic order so that we can use binary search to find the lower and upper bound of arc indices. 

The n-best hypotheses from first pass are rescored in the second pass using a word-based neural language model (NLM) trained on in-domain and external text data, and contain a word embedding matrix of dimension 512 and two LSTM layers of dimension 512 hidden units, with a vocabulary size of 240k words. The model also contains a tuned word reward which penalizes shorter utterances. The n-gram LMs built for n-gram selection and boosting FST generation are all order 4 LMs and are trained with Katz smoothing on and pruned to get the smallest overall model size without compromising the accuracy. Both $LM_{OOD}$  and $LM_{GEN}$ are trained on millions of utterances. 
\begin{table}[ht]
\centering
\begin{tabular}{lcc}
Testset      & RNNT & RNNT + Boosting FP LM \\ \hline
ArtistName   & -    & 6.2                \\
AlbumName    & -    & 11.8               \\
SongName     & -    & 9.1                \\
ItemName     & -    & 8.0                \\
Control Testset & -    & 0.5               
\end{tabular}
\caption{n-best Oracle WERR improvements (n=8) (in \%, +ve numbers means improvement)}
\label{table3}
\end{table}

\begin{table*}[]
\centering
\begin{tabular}{lccccc}
Testset                                & ArtistName & AlbumName & SongName & ItemName & Control Testset \\ \hline
RNNT                                   & -          & -         & -        & -        & -            \\
RNNT + adapted LM FP         & 0.5     & 1.7    & 0.6   & 1.0   & 0          \\
RNNT + boosting LM FP                   & 6.5     & 12.2   & 11.5  & 8.4   & -0.2      \\ \hline
RNNT + SP adapted NLM                  & 9.2     & 11.5   & 9.1   & 15     & 4          \\
RNNT + boosting LM FP + SP adapted NLM & 14.8    & 20.9   & 16.1  & 21.3  & 5         
\end{tabular}
\caption{Additive Improvements with second-pass adaptation (WERR in \% +ve numbers means improvement)}      
\label{table2}
\end{table*}


\begin{table*}[ht]
\centering
\setlength{\tabcolsep}{7pt}
\renewcommand{\arraystretch}{1.25}
\begin{tabular}{lcccccccc}
             & \multicolumn{1}{l}{} & \multicolumn{3}{c}{$T=2$} & \multicolumn{2}{c}{$T=2.5$} & $T=3$  & \multicolumn{1}{l}{} \\ \cline{3-8}
Dataset      & Baseline          & $\lambda=0.25$ & $\lambda=0.5$ & $\lambda=0.75$ & $\lambda=0.25$       & $\lambda=0.5$       & $\lambda=0.25$ & \#utts               \\ \hline
OOD1         & -                    & -59  & -0.3  & -2.2   & 0.3          & 0.2         & 0.4    & 24644                \\
OOD2         & -                    & 24.6   & 32.7  & 34.8   & 24.1         & 32.4        & 23.9   & 827                  \\
OOD3         & -                    & -11.8  & -1.2  & -4.2   & -2.8         & -0.4        & -2.6   & 621                  \\
OOD4         & -                    & -23.1  & 8.1   & -0.6    & 4.9          & 4.9         & 3.3    & 497                  \\
OOD5         & -                    & 4.0    & 28.4  & 21.6   & 25.9         & 25.9        & 24.6   & 497                  \\
OOD6         & -                    & 16.6    & 27.0  & 28.8   & 16.3         & 22.8        & 16.7   & 601                  \\
OOD7         & -                    & -105.8 & 37.7  & 37.1   & 24.2          & 33.1        & 24.0   & 5236                 \\
OOD8         & -                    & -2.6   & 16.5  & 15.1   & 11.3         & 14.0        & 11.1    & 4684                 \\
OOD9        & -                    & -48.6  & -0.8  & -5.5   & 2.0          & -0.5        & 1.1    & 2038                 \\
All OODs Combined        & -                    & - & -  & -   & -          & -        & 2.6   & 599882                 \\
Control Testset   & -                    & -0.7   & -2.0  & -5.0   & -0.3         & -1.5        & -0.3   & 157872              
\end{tabular}
\caption{Impact of threshold and boosting weight (WERR in \% +ve numbers means improvement)}
\label{table1}
\end{table*}

\section{Datasets and Results}
\label{sec:subsubhead}

\subsection{Comparison to LM adaptation}
We evaluate the effectiveness of the proposed adaptation method compared to standard language model adaptation. As mentioned in section ~\ref{sec:methodology}, our goal is to not completely remove the internal LM and hence we don't evaluate against internal LM subtraction or density ratio approaches. Instead, we compare the proposed non-stochastic boosting approach against an adapted n-gram model when used for shallow fusion with the RNN-T model.   

Here, we focus on a dataset with entity names - ItemName, ArtistName, AlbumName and SongName. We use large catalogues to generate synthetic training data using carrier phrases like ``buy [\#ItemName]" or ``play [\#ArtistName]". The carrier phrases are generated from annotated general purpose traffic utterances. For evaluating the model performance, we generated synthetic audio utterance datasets by using AWS Polly\cite{polly}. Row 1, 2 and 3 in Table \ref{table2} compare model performance without second pass rescoring. Row 2 shows that WERR obtained when using a large conventional n-gram SLM adapted to OOD data via constrained optimization \cite{gandhe2018scalable} provide small WERR over Baseline RNN-T. In contrast, Row 3 shows WERR from using boosting style contextual LM (with LLR clipping threshold of 6.5) at a weight of 0.2 in shallow fusion with first pass of RNN-T . The model gives 6\% - 12\% WERR and 6\% - 12\% n-best Oracle WERR at depth 8 (Table \ref{table3}) on various testsets. We discuss how the operating point is chosen in the section ~\ref{sec:operating_point}.

Next, we use an NLM adapted towards the entities for second pass (SP) rescoring. The NLM is trained on a mix of synthetically generated dataset and general dataset and each batch of training gets 15\% synthetic data and the rest from general data. The second pass rescoring alone shows 9\% - 15\% improvement. However, when both the first pass boosting LM and second pass rescoring NLM are used (Table \ref{table2} Row 5), we see maximum improvements, (15\% - 21\%). The gains from first pass and second pass adaptation are additive because the boosting FST gives significant n-best Oracle WERR (refer Table \ref{table3}), which is harnessed by second pass rescoring NLM.

\subsection{Impact of threshold and boosting weight}
\label{sec:operating_point}
To assess the scalability of the proposed solution, we adapt the RNN-T model to large number of domains (144). We use training datasets with 500 - 25000 transcriptions per domain and build a single boosting FST for them as described in section ~\ref{sec:methodology}. We test them using corresponding transcribed utterance testsets. We also evaluate on a control testset (CT) which is created using general traffic (160k utterances).

Our method relies on two hyper-parameters - the threshold $T$ for pruning the n-grams to be boosted, and the SF boosting weight $\lambda$ in Eq.~\ref{eq:eq_sf}. Table ~\ref{table1} shows the results on a select subset of all domain testsets (represented by OOD1  to OOD9) for different values of $\lambda$ and $T$. The datasets with highest measured baseline WER were selected for operating point search as they are most likely to have rare words and would show the most impact. The best results are observed at clipping threshold $3$ and boosting weight $0.25$. 8 out of 9 testsets show significant WER improvement without degradation on general traffic testset. At lower thresholds, the number of target domain n-grams to be boosted increase, which improves the OOD testset WER but it leads to errors in the control testset. We choose an operating point which gave maximum improvement on the OOD testset with \textless{} 0.5 \% WER degradation on the control testset. While the micro-average improvement for all domains was 2.6\%, out of 144 OOD testsets, 74 showed WERR greater than 5\% and 56 showed WERR greater than 10\%. Some of the largest improvements observed in the Table \ref{table1} are in OOD8 (feature for playing sports team matches), OOD5 (feature for grocery checkout) and OOD2 (feature for pay for gas). The domain for OOD8 testset contains rare words like ``freiburg", ``borussia", ``stutgart", which are rare in the general dataset. Though the phrases in domains of OOD2 and OOD5 do not contain rare words, they have n-grams with very different distribution from the general traffic which is heavily biased towards requests from popular domains such as music, informational, smart devices etc. 

\section{Conclusion}
\label{sec:conclusion}
In this paper, we presented a simple and effective method to bias an E2E ASR model output towards new domains using a boosting n-gram FST in shallow fusion with an RNN-T model for ASR. We get about 10\% improvement in 1-best WER and Oracle WER on target domain, without causing any degradation on the general dataset. We also show that these improvements are complementary to the improvements from second pass rescoring methods to obtain additive overall performance gains. In the future, we will evaluate the effectiveness of boosting with other E2E models (such as attention encoder-decoder), as well as other complementary adaptation techniques (like internal language model subtraction).





\bibliographystyle{IEEEbib}
{\small
\bibliography{strings,refs}}

\end{document}